\documentclass[letterpaper, 10 pt, conference]{ieeeconf}
\IEEEoverridecommandlockouts                              % This command is only needed if 
\usepackage{cite}
\usepackage{multirow}
\usepackage{algorithm}      
\usepackage{algorithmic}
\usepackage{graphicx}
\usepackage{times}
\usepackage{amsmath}
\usepackage{amssymb}
\usepackage{subfigure}
\usepackage{graphics} % for pdf, bitmapped graphics files
\usepackage[utf8]{inputenc}
\usepackage{boldline}
\usepackage{multicol}
\usepackage{mathtools}
\DeclarePairedDelimiter{\norm}{\lVert}{\rVert}

\usepackage[pdfstartview=FitH,bookmarksnumbered=true,colorlinks,bookmarksopen=true]{hyperref}
\title{\LARGE \bf
    A Mobile Robot Hand-Arm Teleoperation System by Vision and IMU
}

\author{
    Shuang Li$^{1}$, Jiaxi Jiang$^{2}$, Philipp Ruppel$^{1}$, Hongzhuo Liang$^{1}$, Xiaojian Ma$^{3}$, \\ Norman Hendrich$^{1}$,
    Fuchun Sun$^{4}$,
    Jianwei Zhang$^{1}$
    \thanks{$^{1}$TAMS (Technical Aspects of Multimodal Systems), Department of Informatics, Universit\"{a}t Hamburg}
    \thanks{$^{2}$Department of Computer Science, RWTH Aachen University}
    \thanks{$^{3}$Center for Vision, Cognition, Learning and Autonomy, Department of Statistics, University of California, Los Angeles}
    \thanks{$^{4}$Beijing National Research Center for Information Science and Technology (BNRist), State Key Lab on Intelligent Technology and Systems, Department of Computer Science and Technology, Tsinghua University}
}
\begin{document}
    
    \maketitle
    \thispagestyle{empty}
    \pagestyle{empty}
    
    %%%%%%%%%%%%%%%%%%%%%%%%%%%%%%%%%%%%%%%%%%%%%%%%%%%%%%%%%%%%%%%%%%%%%%%%%%%%%%%%
    \begin{abstract} %0.2-0.4page
    In this paper, we present a multimodal mobile teleoperation system that consists of a novel vision-based hand pose regression network (Transteleop) and an IMU-based arm tracking method. 
    Transteleop observes the human hand through a low-cost depth camera and generates not only joint angles but also depth images of paired robot hand poses through an image-to-image translation process.
    A keypoint-based reconstruction loss explores the resemblance in appearance and anatomy between human and robotic hands and enriches the local features of reconstructed images.
    A wearable camera holder enables simultaneous hand-arm control and facilitates the mobility of the whole teleoperation system.
    Network evaluation results on a test dataset and a variety of complex manipulation tasks that go beyond simple pick-and-place operations show the efficiency and stability of our multimodal teleoperation system.
    \end{abstract}
    
    \section{Introduction} %0.6-1page
    Teleoperation is a crucial research direction in robotics with many applications such as space, rescue, medical surgery, and imitation learning \cite{argall2009survey}. 
    And it is still superior to intelligent programming when it comes to making fast decisions and dealing with corner cases.
    However, teleoperation of an anthropomorphic robotic hand to perform dexterous manipulation is still challenging. 
    Markerless vision-based teleoperation offers strong advantages as it is low-cost and less invasive.
    For instance, Dexpilot~\cite{dexpilot} recently demonstrated the abilities of vision-based methods which can compete with other teleoperation methods using tactile feedback. 
    Since the robot hand and the human hand occupy two different domains, how to compensate for kinematic differences between them plays an essential role in markerless vision-based teleoperation. 

    To tackle this issue, we introduce an image-to-image translation~\cite{pix2pix, bicyclegan, pathak2016context} concept for vision-based teleoperation methods.
    Image-to-image translation, which aims to map a representation of a scene into another, is also a prevalent research topic widely used in collection style transfer, object transfiguration, and imitation learning.
    The key to image-to-image translation is to discover the hidden mapping feature between two representations.
    In our case, we would like to find a method that can thoroughly comprehend the kinematic similarity between the human hand and the robot hand.
    We assume that if a robot could translate the observed scene (such as the human hand) to its scene (such as the robot hand), the robot would have perceived valuable hidden embeddings that represent the resemblance of pose features between two image domains.

    \begin{figure}[!t]
        \vskip 0.12in
        \includegraphics[width=0.485\textwidth]{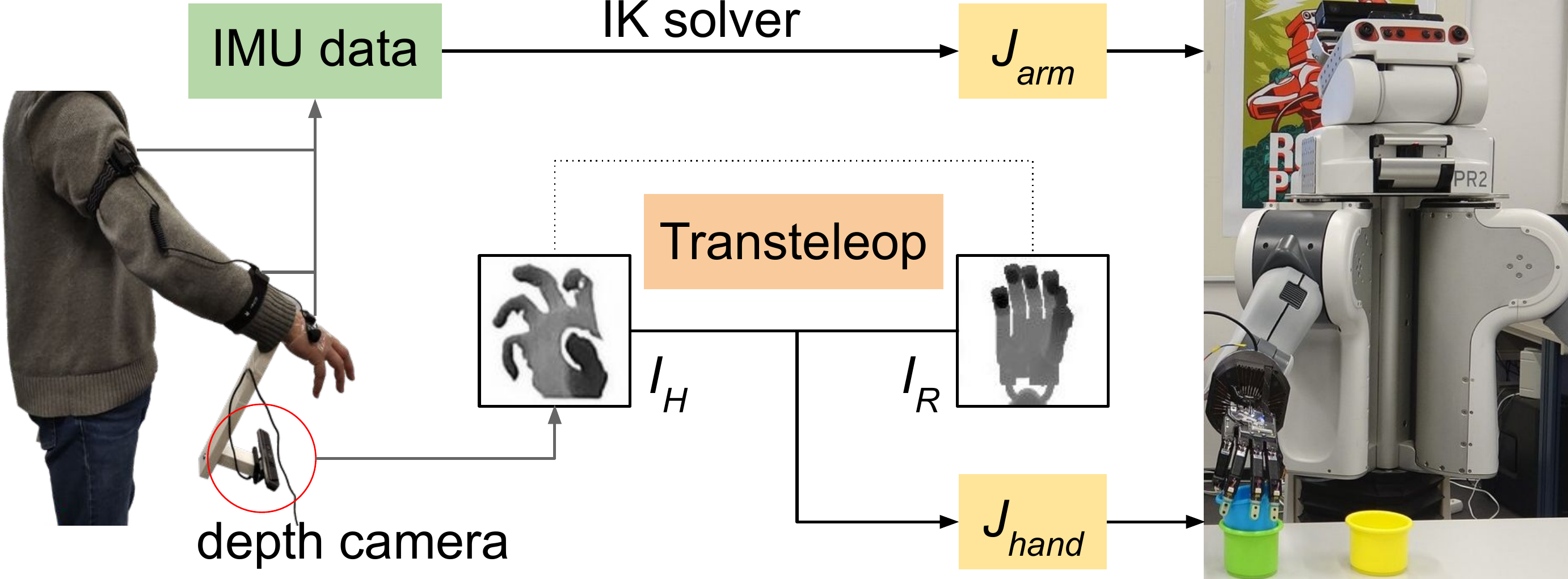}    
        \caption{Our multimodal teleoperation system is built on a vision-based method, Transteleop, which predicts the joint angles of an anthropomorphic hand, and an IMU-based arm control method.
        Transteleop gains meaningful pose information from depth images of a human hand $I_{H}$ based on an image-to-image translation structure and predicts the joint angles $J_{hand}$ of the robot hand.
        Through this multimodal system, we implement different manipulation tasks such as pick and place, cup insertion, object pushing, and dual-arm handover tasks on a PR2 robot with a Shadow hand installed on its right arm.
        }
        \label{framework}
        \vskip -0.15in
    \end{figure}

    To this end, we propose a novel vision-based teleoperation method called Transteleop, which extracts coherent pose features between the paired human and robot hand based on image-to-image translation methods. 
    Transteleop takes the depth image of the human hand as input, then estimates the joint angles of the robot hand, and also generates the reconstructed image of the robot hand. 
    In the spirit of supervised learning, to enhance the richness of the features extracted from the image translation structure, we design a keypoint-based reconstruction loss to focus on the local reconstruction quality around the keypoints of the hand.

    One problem of current vision-based methods is that the hand of the teleoperator should stay inside the limited view range of the camera system. This restriction impedes the operators from completing manipulation tasks that need a wide working area.
    To achieve a truly mobile hand-arm teleoperation system, we develop a camera holder to mount the camera on the arm of a human.
    Additionally, we implement IMU-based teleoperation to control the movements of the robot arm.

    In conjunction with the Transteleop method, a camera holder, and IMU-based arm teleoperation, this multimodal system can not only maintain the natural motion of the human fingers but also allow for flexible arm motion.
    Fig.~\ref{framework} illustrates the framework of our proposed method for hand-arm teleoperation.
    
    The main contributions are summarized below:
    
    1. We set up a novel and multimodal hand-arm teleoperation system for a PR2 Robot equipped with a 19 DoF Shadow hand.
    
    2. A novel vision-based deep learning method, Transteleop, is proposed to estimate the joint angles of the robot and to bridge the kinematic disparities between the robot hand and the human hand. The idea is inspired by image-to-image translation methods.
    
    3. Thanks to a self-designed camera holder, the teleoperator is not limited to a fixed workspace anymore.
    
    4. The demonstration of the teleoperation system across three trained human demonstrators on several subtle and dexterous tasks suggests the reliability of our system.
    
\section{Related Work}%0.8-1page
    %\noindent
    %\textbf{Markerless vision-based teleoperation.}
    \subsection{Markerless vision-based teleoperation}
    Markerless vision-based teleoperation~\cite{tele2007advances, du2013markless} of robots offers the advantages of allowing for natural, unrestricted limb motions and of being less invasive. 
    Especially markerless methods are suited to dexterous teleoperation, which requires capturing all the essence of finger motions.
    With the rapid expansion of deep learning methods, 
   leveraging image processing algorithms like hand pose estimation or object segmentation is becoming a new trend in the robotic community.
    Michel \textit{et al.}~\cite{markerless_humanpose} tracked human body motion from markerless visual observations, then mapped human motions to a NAO humanoid robot by the inverse kinematics process.
    Antotsiou \textit{et al.}~\cite{antotsiou2018task} proposed a hand pose estimator and a task-oriented retargeting method to achieve the teleoperation of an anthropomorphic hand in simulation.
    Similar to~\cite{gomez2019accurate}, they put one neural network, HandLocNet, in charge of detecting the hand in the RGB image and used the network HandPoseNet to accurately infer the three-dimensional position of the joints, retrieving the full hand pose.
    Nevertheless, these methods strongly depend on the accuracy of the hand pose estimation or the classification and spend much time on post-processing.
    In our previous work~\cite{teachnet}, we proposed the end-to-end neural network TeachNet, which was used with a consistency loss function to control a five-fingered robotic hand based on simulated data. Although this method was efficient, we only demonstrated this method by simplistic grasping and did not show a high level of dexterity.
    As TeachNet only focused on finger motions, all experiments were employed by a robot hand fixed in the same position.
    The learning models in~\cite{dexpilot} provided hand pose and joint angle priors using a fused input point cloud coming from four cameras at fixed positions. Although the teleoperation results were quite impressive, the two-phased data collection procedure they used was hard to replicate. In addition, this system cannot carry out manipulation tasks that require a large motion range because of the restricted workspace of the pilot.

    %\noindent
    %\textbf{Wearable device-based teleoperation.}
    \subsection{Wearable device-based teleoperation}
    Robotic teleoperation has usually been implemented through different types of wearable devices such as marker-based tracking~\cite{SCHUNKS5FHteleop}, inertial measurement units (IMU)~\cite{fang20183d}, electromyography (EMG) signal sensors~\cite{luo2019teleoperation}, virtual/mixed reality devices~\cite{krupke2018comparison} and highly promising haptic devices~\cite{haptx}. 
    Regarding dexterous teleoperation, glove-based methods must be customized and easily obstruct natural joint motions, while IMU-based and EMG-based methods provide less versatility and dexterity.
    On the other hand, regarding teleoperation of a multiple degree of freedom (DoF) robotic arm, contact-based methods are convenient to implement and efficient enough.
    For instance, Fang \textit{et al.}~\cite{fang2015robotic} established a multimodal fusion algorithm of a self-designed IMU device to deduce the orientations and positions of the human arm and hand. 
    The robot experiments demonstrated a smooth arm control of the UR5 robot and the SCHUNK arm but a shallow grasping and releasing ability of the Barrett robotic hand.
    Motion capture systems provided accurate tracking data but can be expensive, and the correspondence problem between markers on the fingers and cameras still needs to be solved.
    Zhang \textit{et al.}~\cite{zhuang2019shared} presented a motion capture system and a myoelectric device for teleoperative control over a hand-arm system. The system worked convincingly well as a prosthetic HMI, but experiments did not reveal how well the system works for intricate in-hand manipulation tasks.
    Besides that, tactile feedback is essential in contact-rich dexterous manipulation tasks~\cite{johansson2009coding, jain2019learning}.
    Haptic devices are widely investigated in surgical robots and are used to collect training data in virtual reality applications~\cite{broeren2004virtual}.

    In this paper, to attain natural human finger motion, we will further explore a markerless method to discover appearance and anatomy of human and robotic hands and to verify the performance of our approach regarding dexterous manipulation.
    To create a complete robotic hand-arm system, we will develop an arm tracking system using an IMU setup, which is easy to implement and suitable to achieve accurate control.

    %\noindent
    %\textbf{Image-to-image translation methods.}
    \subsection{Image-to-image translation methods}
    Image-to-image translation has been widely researched on different generative models such as restricted Boltzmann machines~\cite{CarreiraPerpin2005OnCD}, auto-encoder models~\cite{kingma2013auto}, generative adversarial networks (GANs)~\cite{gan}, and several variants of these models. 
    In the robotic field, image-to-image translation methods have been employed to map representations from humans to robots. 
    Simith \textit{et al.}~\cite{smith2019avid} converted the human demonstration into a video of a robot and generated image instructions of each task stage by performing pixel-level image translation.
    They constructed a reward function for a reinforcement learning algorithm through translated instructions and evaluated the proposed method in a coffee machine operation task.
    Sharma \textit{et al.}~\cite{third} decomposed third-person imitation learning into a goal generation module and an inverse control module. 
    The goal generation module translated observed goal states from a third-person view to contexts of the robot by translating changes in the human demonstration images.
    All the above methods indicate that image-to-image translation methods are capable of learning shared features between mapping pairings.
    Hence, we will study how to use this method to extract shared features between depth images of the human hand and the robot hand.
    
    \begin{figure*}[!ht]
        \includegraphics[width=0.735\textheight]{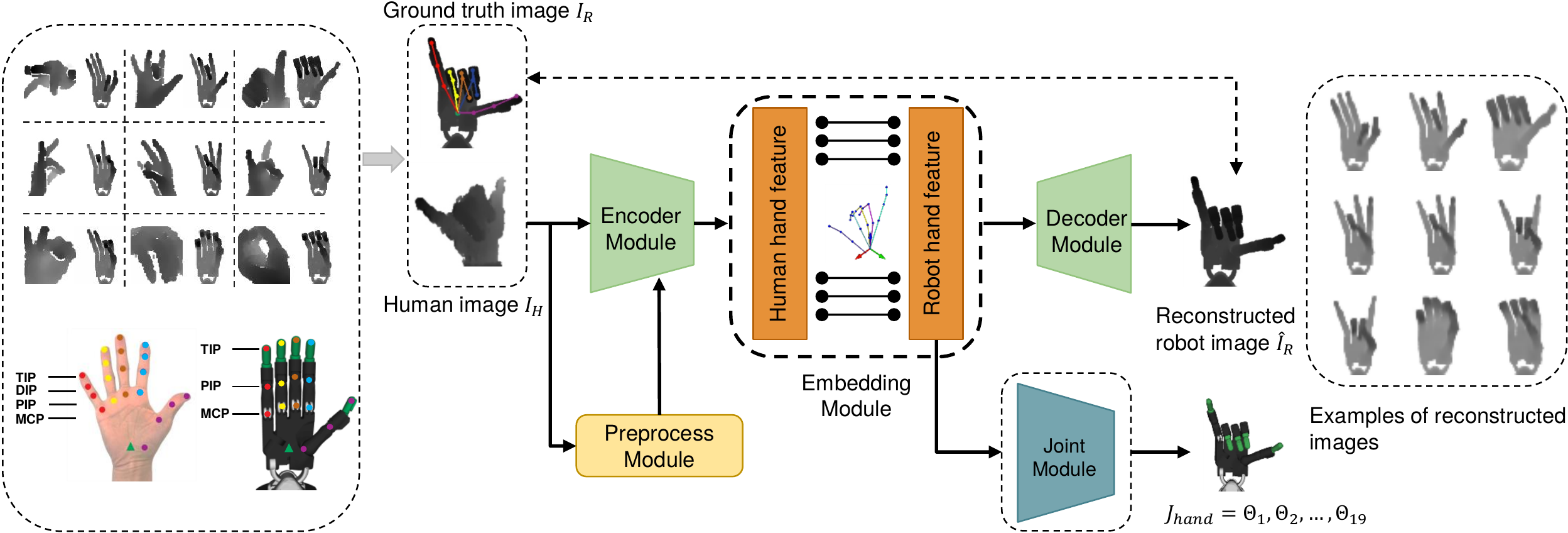}    
        \caption{Transteleop Architecture.
            Left: Examples of paired human-robot hand datasets used to train Transteleop. 
            Center: The encoder-decoder structure is an image-to-image translation branch, which feeds depth images of a human hand $I_H$ and produces reconstructed depth images of the robot hand $I_R$.
            The joint module takes the pose embedding from the encoder-decoder structure and predicts the robot's joint angles $J_{hand}$.
            The preprocess module is a spatial transformer network which explicitly permits the spatial manipulation of input images.
        Right: Examples of reconstructed robot images are generated from Transteleop, whose inputs are the dataset examples shown on the left.
        }
        \label{net}
        \vskip -0.15in
    \end{figure*}
    
    \section{Problem Formulation}
    Our goal is to build a mobile robotic hand-arm teleoperation system in which the teleoperator is in an unlimited workspace and performs natural hand motion for a series of manipulation tasks.
    To set up such a system, we formulate a novel vision-based method to teleoperate the anthropomorphic hand and utilize an IMU-based device to control the arm simultaneously.
    Let $I_H$ be the image of a human demonstrating hand poses of manipulation tasks as observed by a depth camera. 
    The vision part aims to train a neural model that feeds $I_H$ and predicts joint angles $J_{hand}$ of the robot while the IMU part intends to map the absolute motion of the human arm to the robot arm.
    The robot system used in this work is a PR2 robot with a 19 DoF Shadow hand mounted on its right arm, as shown in Fig.~\ref{framework}. 
    Unlike the 7 DoF left arm of PR2, the right arm of PR2 only has 5 DoF due to the attached Shadow hand.
    
    \section{Transteleop: Vision-based Teleoperation by Image-to-image Translation} %1.5-2pages
    We propose a novel network Transteleop to discover the kinematic resemblance between the human hand and the robot hand and to predict the joint angles of the robot hand using the image-to-image translation method.
    
    \subsection{Transteleop} %0.2-0.5pages
    Imagine that we have an image $I_R$ of a robotic hand from a fixed viewpoint and an image $I_H$ of a human hand in random global orientation, while the robotic hand in the image acts exactly the same as the human hand. 
    Even though the bone length, the global pose, and the joint range of these paired hands are distinct, the pose feature $Z_{pose}$ such as the skeletal shape and the whole silhouette will reveal the similarity between them.
    We believe that it would be very favorable to predict $J_{hand}$ from the shared pose feature $Z_{pose}$ rather than the bare $I_H$.
    In order to attain an instructive feature representation $Z_{pose}$, we adopt a generative structure that maps from image $I_H$ to image $I_R$ and retrieves the pose from the bottleneck layer of this structure as $Z_{pose}$. 
    Although conditional GANs have led to a substantial boost in the quality of image generation, the discriminator pursues the high reality
    %and low blurriness 
    of reconstructed images but does not fully concentrate on the pose feature of the input. For this reason, we choose an auto-encoder structure to sustain our pipeline, as visualized in Fig.~\ref{net}.
    
    \noindent
    \textbf{Encoder-decoder module}. 
    The encoder takes a depth image of a human hand $I_H$ from various viewpoints and discovers the latent feature embedding $Z_{pose}$ between the human hand and the robot hand. We use six convolutional layers containing four downsampling layers and two residual blocks with the same output dimension. Thus, given an input image of size $96 \times 96$, the encoder computes an abstract $6 \times 6 \times 512$ dimensional feature representation $Z_{H}$.
    
    Similar to~\cite{pathak2016context}, we connect the encoder and the decoder through fully-connected layers instead of convolutional layers because the pixel areas in $I_H$ and $I_R$ in our dataset are not matched.
    This design results from the fact that a fully-connected layer allows each unit in the decoder to reason on the entire image content. In contrast, a convolutional layer cannot directly connect all locations within a feature map.  
    Through the embedding module, we extract the useful feature embedding $Z_{pose}$ from the 8192-dimensional feature representation $Z_{H}$.
    
    The decoder aims to reconstruct a depth image of the robot hand $\hat{I}_R$ from a fixed viewpoint from the latent pose feature $Z_{pose}$. 
    One fully-connected layer connects the feature from the $Z_{pose}$ to robot feature vector $Z_R$.
    Four up-convolutional layers with learned filters and one convolutional layer for image generation follow.
    
    Unlike common image-to-image translation tasks, the generated image $\hat{I}_R$ should care more about the accuracy of local features such as the position of fingertips instead of global features such as image style. 
    This is because the pixels of the joint keypoints possess more information about the hand pose.
    Regarding the Shadow hand, as depicted in Fig.~\ref{net}, each finger has three keypoints. 
    Therefore, we designed a keypoint-based reconstruction loss to capture the overall structure of the hand and concentrate on the pixels around the 15 keypoints of the hand.
    The scaling factor of each pixel error is determined by how close this pixel is to all keypoints.
    We regard the eight neighboring pixels of each keypoint as important as these keypoints themselves.
    The reconstruction loss $L_{recon}$ is an L2 loss that prefers to minimize the mean pixel-wise error but does not encourage less blurring, defined as:    
    \begin{align}\label{recon_loss}
    \mathcal{L}_{recon} = 
    \frac{1}{N}\sum_{i=1}^{N} \alpha_i \cdot (I_{R,i} - \hat{I}_{R,i})^2
    \end{align}
    where $N$ is the number of pixels and $\alpha_i \in [0, 1]$ is the weighting factor of the i-th pixel. 
    
    \begin{figure}[!ht]
        \centering
        \includegraphics[width=0.38\textwidth]{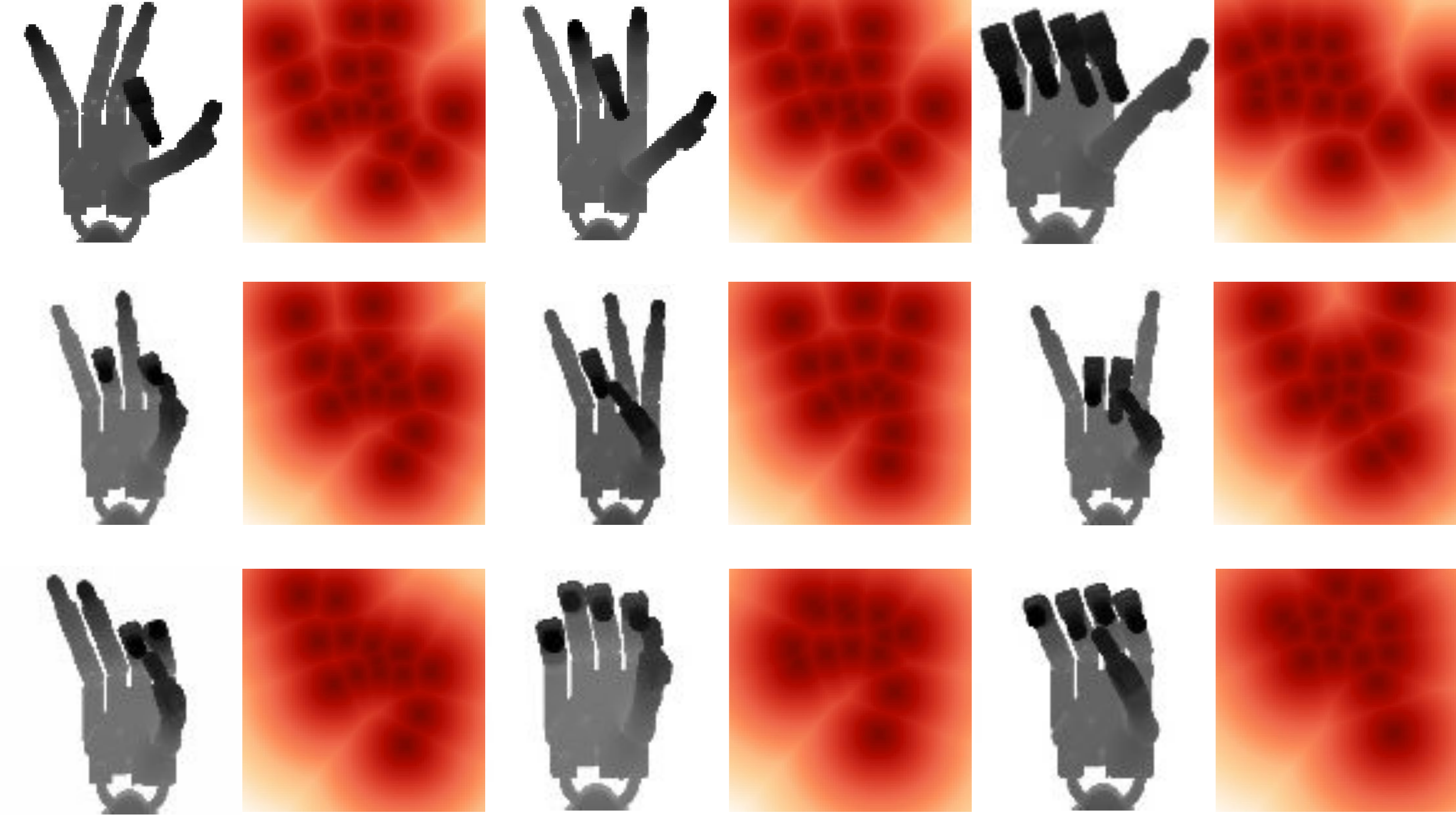}    
        \caption{The heatmap of scaling matrix $\alpha$. The darker color illustrates how important these pixels are.}
        \label{weight}
        \vskip -0.08in
    \end{figure}
        
    \noindent
    \textbf{Joint module}. 
    The joint module focuses on deducing 19-dimensional joint angles $J_{hand}$ from the latent feature embedding of the decoder.
    We choose $Z_R$ instead of $Z_{pose}$ because $Z_R$ has richer features depicting the pose feature of the robot hand.
    Three fully-connected layers are employed in the joint module.
    The joint module is supervised with a mean squared error (MSE) loss $\mathcal{L}_{joint}$
    \begin{equation}
    \label{angloss}
    \mathcal{L}_{joint} = \frac{1}{M}\norm{J_{hand} - J}^2_{2}
    \end{equation}
    where $M$ is the number of joints and $J$ denotes the ground truth joint angles.
    
    Overall, the complete training objective:
    \begin{equation}
    \label{wholeloss}
    \mathcal{L}_{hand} = \lambda_{recon} \cdot \mathcal{L}_{recon} + \lambda_{joint} \cdot \mathcal{L}_{joint}
    \end{equation}
     where $\lambda_{recon}, \lambda_{joint}$ are the scaling factors.

    \subsection{Paired dataset} %0.2-0.3pages
    \label{dataset}
    We trained Transteleop based on a recently released dataset of paired human-robot images from~\cite{teachnet}. This dataset contains 400K pairs of simulated robot depth images and human hand depth images. 
    The ground truth are 19 joint angles of the robot hand, which correspond to the middle, proximal, and the metacarpal joints of five fingers plus one additional joint for the little finger and the thumb, respectively.
    This paired dataset records nine depth images of the robot hand from different viewpoints simultaneously, corresponding to one human pose. Considering to abstract an explicit kinematic configuration from the robot image, we only use the robot images taken in front of the robot, as shown on the left side of Fig.~\ref{net}.
    
    One challenge of training Transteleop is that the poses of 
    the human hand vary considerably in their global orientations.
    Thus, we applied a spatial transformer network (STN)~\cite{STN},
    which provides spatial transformation capabilities of input images, before the encoder module. Due to the invariance of the outputs $J_R$ and $I_R$ to the spatial transformation of the input image, we do not need to modify the ground truth.
    
    \section{Hand-arm Teleoperation System}  % (0.2-0.5 pages)
\begin{figure}[!ht]
        \centering
        \includegraphics[width=0.2\textheight]{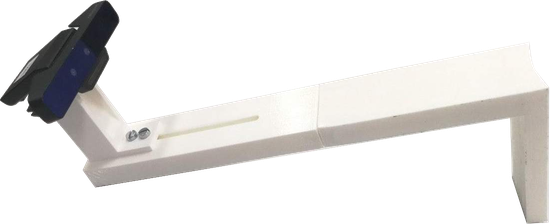}    
        \caption{The camera holder is used to mount the camera on the human arm.
        }
        \label{holder}
    \end{figure}

    However, the hand of the teleoperator easily disappears from the field of view of the camera if the arm movement is relatively large. 
    Even though a multicamera system could be one of the solutions for this, we solve this problem by a cheap 3D-printed camera holder, which can be mounted on the forearm of the teleoperator, as shown in Fig.~\ref{holder}. Consequently, the camera will move along with the arm. The whole weight of the camera holder is 248\,g, and the camera used in our experiments is the Intel RealSense SR300 depth sensor, whose weight is 108\,g. 

    Due to the uncertainty of the camera position, we use the Perception Neuron (PN) device~\cite{pn} to control the arm of the robot, and thus extend this teleoperation system to the hand-arm system. Perception Neuron is an IMU-based motion capture device that is popular in the fields of body tracking and VR / Game interaction.
    Regarding single-arm tracking, three IMU elements are sufficient to capture the motion of the palm, upper arm, and forearm.
    We set the global frame of PN to be parallel to the robot base frame.
    Depending on the rotation data from PN and the link length of the robot arm, we calculate the wrist pose of the robot. Then we compute the joint angles of the robot arm by feeding this pose to the BioIK solver~\cite{bioik}. After this, we set the angular velocity $\mathcal{V}_{t}$ of each joint by calculating and scaling the feedforward joint difference between the desired joint angles of the current frame $J_t^{ik}$ and of the previous frame $J_{t-1}^{ik}$ and the feedback joint difference between the desired joint angle of the current frame $J_{t}^{ik}$ and of the current robot joint state $J_{t}^{robot}$.
    \begin{align}\label{vel}
    \mathcal{V}_{n,t} = \delta_1\cdot (J_{n,t}^{ik} - J_{n,t-1}^{ik}) + \delta_2 \cdot (J_{n,t}^{ik} - J_{n,t}^{robot})
    \end{align}
    where $n$ is the n-th joint of the arm, $\delta_1, \delta_2$ account for the scaling factor of each velocity term.
    
    \section{Network Evaluation} % (0.8-1 pages)
    \label{sec_net}
    
    \begin{figure*}[!ht]
    \centering
        \label{teleop_combine}
        \includegraphics[width=1\textwidth]{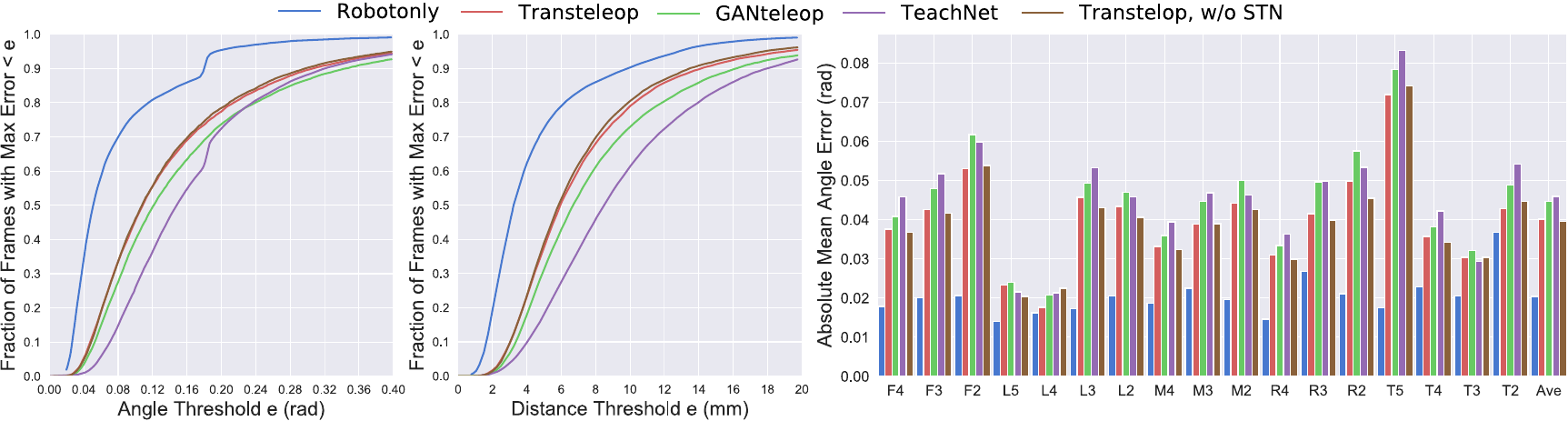}
        { \footnotesize (a)~~~~~~~~~~~~~~~~~~~~~~~~~~~~~~~~~~~~~~~~~~(b)~~~~~~~~~~~~~~~~~~~~~~~~~~~~~~~~~~~~~~~~~~~~~~~~~~~~~~~~~~~~~~~~~(c)~~~~~~~~~~~~~}
        \caption{(a) and (b): The fraction of frames whose absolute maximum joint angle/distance error is below a threshold between the Transteleop approach and different baselines on our test dataset.
        (c): Comparison of the absolute average angle error on the individual joint between the Transteleop approach and different baselines on our test dataset. F means the first finger, L means the little finger, M means the middle finger, R means the ring finger, T means the thumb. 2, 3, 4, 5 mean the n-th joint of the finger.}
        \label{net_eval}
         \vskip -0.15in
    \end{figure*}
    
    \subsection{Optimization and inference details}
    The input depth images are extracted from the raw depth image as a fixed-size cube around the hand and resized to $96 \times 96$.
    To optimize our networks, we use minibatch stochastic gradient descent and apply Adam optimizer with a learning rate of 0.002 and momentum parameters $\beta_1 = 0.5, \beta_2 = 0.999$.
    We add a batch normalization (BN) layer and a rectified linear unit (ReLU) after each convolution layer. 
    ReLU is also employed as an activation function after all FC layers except for the last FC layer.
    We use $\lambda_{recon} = 1, \lambda_{joint} = 10$.
    At inference time, we only run the encoder module and the joint module for joint angle regression.
 
    \subsection{Transteleop evaluation}
    \label{sec_joints}
    
    To evaluate whether Transtelop could learn indicative visual representations, 
    we compared the Transtelop method with two network structures: TeachNet~\cite{teachnet} and GANteleop, which adds a PatchGAN discriminator and an adversarial loss based on the ``pix2pix'' framework~\cite{pix2pix}.
    Additionally, to show the regression results from the robot's own domain, we trained a model that removes the decoder module in Transteleop and only feeds the images of the robot hand. This baseline is referred to as Robotonly.
    All images fed into Robotonly are taken from a fixed third-person viewpoint.
    To examine whether the STN module learns invariance to the hand orientation and rotation, we also trained Transtelop without using STN baseline, which has the same structure as Transtelop but without the preprocess module. 
    
    We evaluated the regression performance of Transteleop and four baselines on our test dataset using standard metrics in hand pose estimation: 1) the fraction of frames whose maximum joint angle errors are below a threshold; 2) the fraction of frames whose maximum joint distance errors are below a threshold; 3) the average angle error over all angles in $\Theta$.
    
    In Fig.~\ref{net_eval}, the Robotonly model significantly outperforms other baselines over all evaluation metrics because of the matched domain and the identical viewpoint.
    Transtelop and GANteleop both show an average 10.63\% improvement of the accuracy below a maximum joint angle, which is higher than that of TeachNet.
    We infer that both image-to-image translation methods seize more pose features from the robot than TeachNet because these methods have to generate a whole image of the robot hand.
    Examples of generated robot images by Transteleop are visualized in the right part of Fig.~\ref{net}.
    Even if the reconstructed images are blurry, they are rather similar to the ground truth in the left of Fig.~\ref{net}, which explicitly proves that the pose feature $z_R$ in the embedding module could contain enough pose information.
    Moreover, the reason why GANteleop worse than Transteleop is that the discriminator in GANteleop focuses on pursuing realistic images and weakens the supervision of $\mathcal{L}_{joint}$.
    Comparing Transtelop and Transtelop without using STN, there is no significant improvement due to the STN module. This suggests that the additional spatial transform brings a little appearance normalizing effect to this task, but does not significantly promote the hand pose transform to a canonical pose.
        
    As illustrated in Fig.~\ref{net_eval}(c), the absolute average error of the joint regression of all methods is lower than $0.05$\,rad.
    The highest error happens on thumb joint 5
    because there is a big discrepancy between the human thumb and the Shadow thumb. 

\section{Manipulation Experiments} % (1-1.5 pages)
    The multimodel teleoperation approach was systematically evaluated across four types of physical tasks that analyze precision and power grasps, prehensile and non-prehensile manipulation, and dual-arm handover tasks.
    For the control of the arm, we set $\delta_1 = 0.7, \delta_2 = 0.1$.
    The frequency of the arm's velocity control is 20\,Hz.
    The starting poses of the human arm were always consistent with the starting pose of the robot arm.
    Meanwhile, the arms of the robot always started and ended at almost similar poses over every task.
    The frequency of the hand's trajectory control is set to 10\,Hz.
    One female and two male testers have participated in the following robotic experiments, and each task was randomly performed by one of them. 
    
    \begin{figure*}[!ht]
        \centering
        \subfigure[]
        {\includegraphics[height=0.174\textwidth]{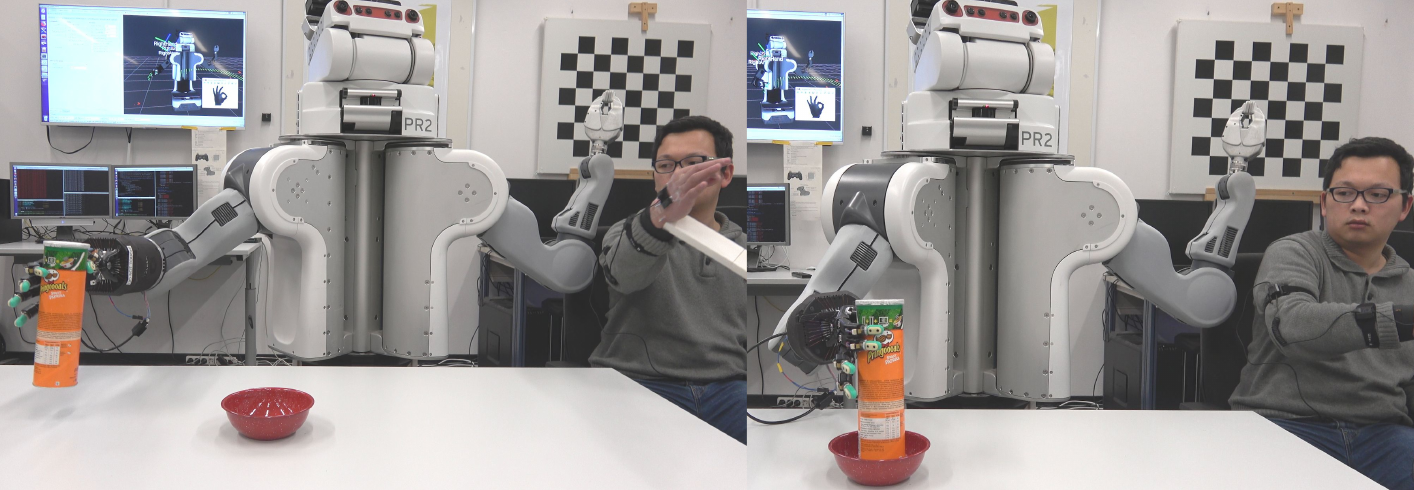}
        \label{pick}}
        \subfigure[]
        {\includegraphics[height=0.174\textwidth]{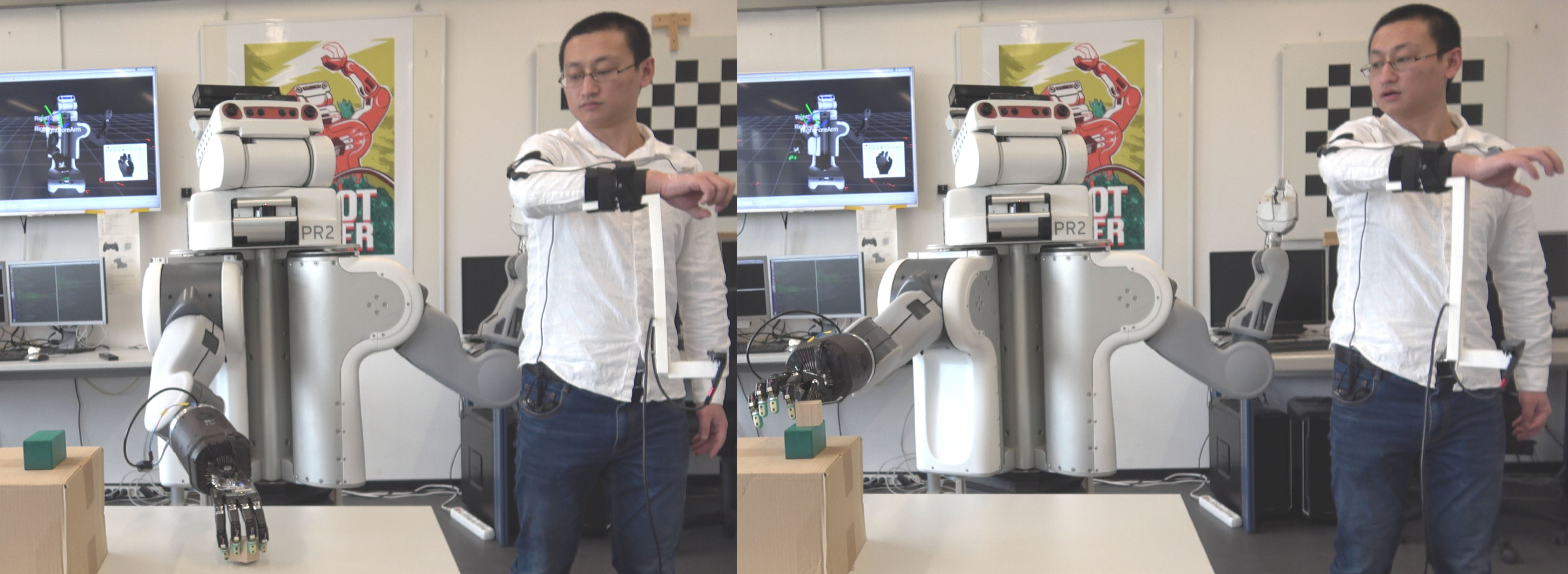}
        \label{pick2}}
        \vskip -2mm
        \caption{\subref{pick} The PR2 robot picks a Pringles can and places it in a bowl. The Pringles and the bowl are set on the same table. \subref{pick2} The PR2 robot picks a wooden cube on the table and places it on a rectangular brick on a box. The height of the box is 325\,mm.
        }
        \vskip -0.1in
        \label{two_picks}
    \end{figure*}
 
    \begin{figure*}[!ht]
        \centering
        \subfigure[]
        {\includegraphics[height=0.1625\textwidth]{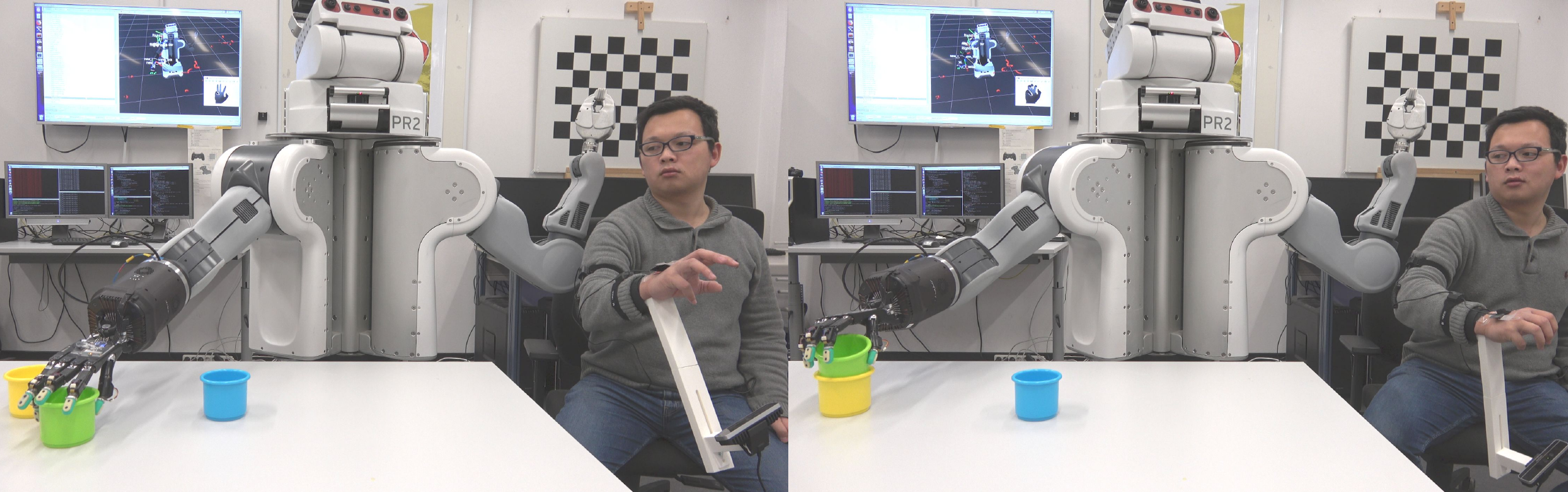}
        \label{cup}}
        \subfigure[]
        {\includegraphics[height=0.1625\textwidth]{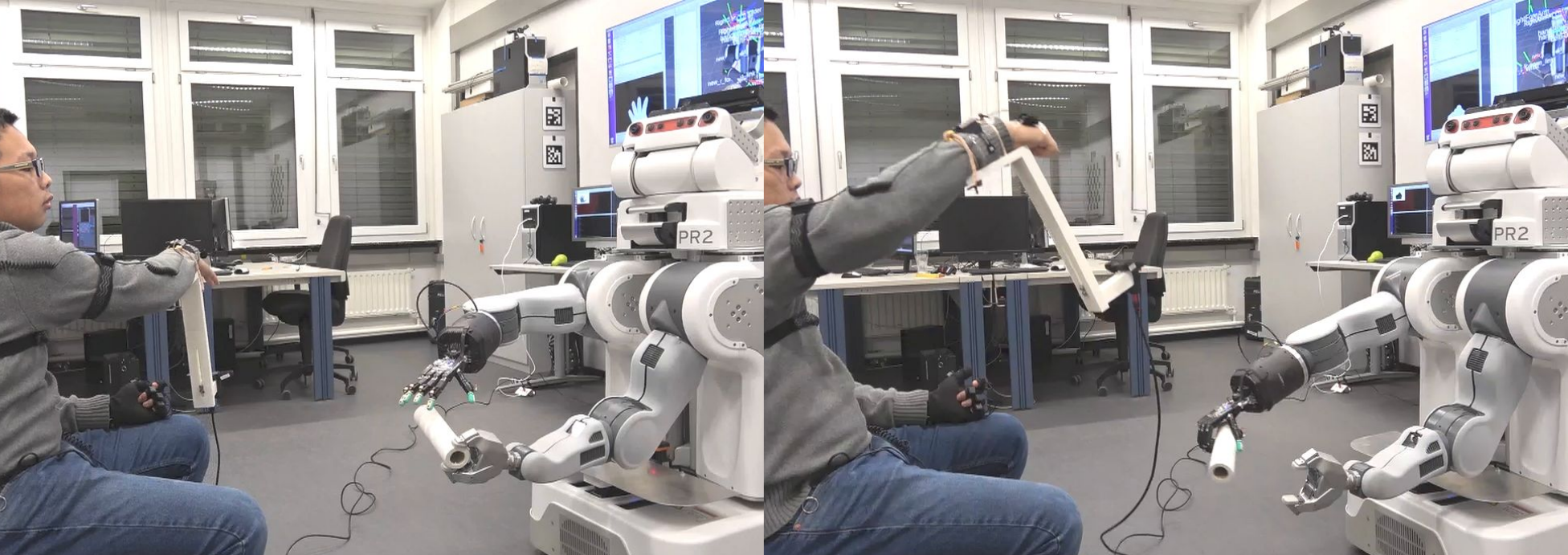}
        \label{handover}}
        \vskip -2mm
        \caption{\subref{cup} The PR2 robot inserts three cups into each other. \subref{handover} The PR2 robot hands a roll of paper from its left gripper over to its right hand.
        }
        \vskip -0.1in
        \label{cup_and_handover}
    \end{figure*}
    
    \begin{figure*}[!ht]
        \centering
        \includegraphics[width=1\textwidth]{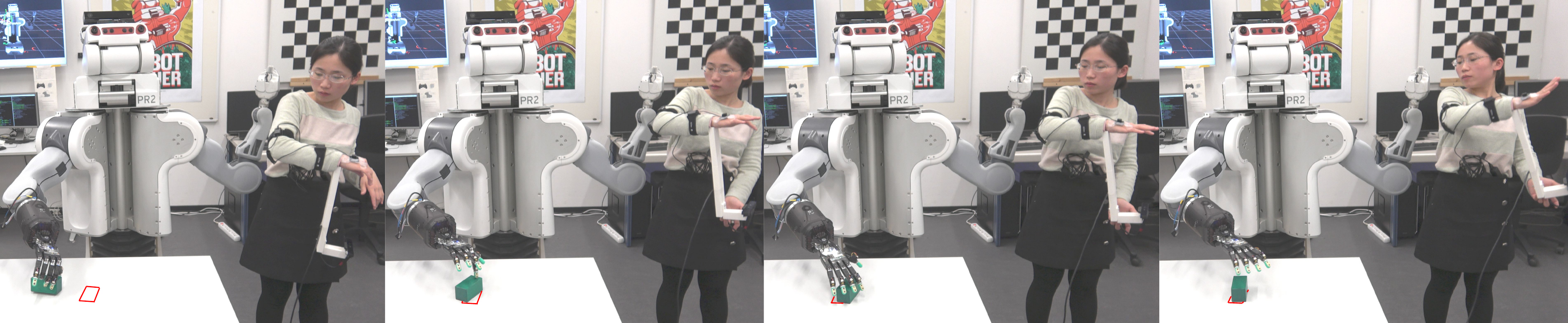}  \vskip -2mm    
        \caption{The PR2 robot pushes a rectangular brick to a specified goal. The red rectangular on the table represents the target pose of the brick.}
        \label{push}
        \vskip -0.1in
    \end{figure*}    
    
    1) Pick and place.
    We prepared two testing scenarios: 
    pick a Pringles can and place it in a red bowl on the same table; pick a cube on the table and place it on top of a brick.
    The first scenario requires the power grasp skills of the robot, and the second scenario needs the precision grasp skills of the robot and a wide-enough workspace for the teleoperator.

    2) Cup Insertion.
   Three concentric cups are to be inserted into each other. This task examines the abilities of precision grasp and releasing.
    
    3) Object pushing. 
    We set random initial poses of a brick then push the brick into a designated pose. This task contains the challenges of pushing, sliding, and precision grasping.
    
    4) Dual-arm handover. The left arm will hand a roll of paper over to the right hand. The operator also exploits the PN setup to control the left arm and left gripper of PR2. This task tests the coordination ability of the whole teleoperation system. Owing to the mobility of our system, the human can sit face to face instead of parallel to the robot to get a clear view of it.
    
    Similar to~\cite{dexpilot}, the operators performed a warm-up training phase for each task with five non-consecutive attempts before the real testing trials.
    For easier tasks such as pick and place, after three trials, the operators could complete the task well.
    But for the handover task, the teleoperator took more trials to adapt to the opposite operating direction of the arm.
    Each task was conducted five times by one demonstrator. 
    The Figs.~\ref{two_picks},~\ref{cup_and_handover},~\ref{push} qualitatively demonstrate the experimental results of our hand-arm system.
    
    \begin{table}[!t]
    \centering
    \caption{Average completion time and success rate of each task}
    \begin{tabular}{ccccccc}\hlineB{2}
                  & pick1 & pick2 & cup & pushing & handover\\ \hline
    Ave. time & 18.5& 37.2& 25.5& 62.0& 36.33\\
    Ave. success rate & 100\% & 100\%& 100\%& 80\%& 60\%\\
    \hlineB{2}
    \end{tabular}
    \label{time}
    \vskip -0.1in
    \end{table}
    
    Table~\ref{time} numerically shows the average completion time a teleoperator took to finish a task, and the success rate.
    The completion time was calculated when the robot started to move until it went back to the starting pose.
    The high success rate and short completion time of two pick and place tasks, and the cup insertion task indicate our system has the ability of precision and power grasps.
    Compared to two pick and place tasks shown in Fig.~\ref{two_picks}, the brick is much smaller than the bowl so that the robot needed a longer time to find a precious place to land the cube.
    During the pushing task, the robot could quickly push the brick close to the target position using multiple fingers.
    Nevertheless, the operator took a long time to deal with the orientation of the brick in order to make the pushing error lower than 5\,mm. 
    The handover task achieved a relatively low success rate, mainly because of the imprecise control of the left gripper, so the robot accidentally lost the object. These results reflect the fact that the visual-based method is more suitable for multi-finger control than the IMU-based method.

\section{Conclusions and Future Work} % (0.25-0.5 page)holistically
    This paper presents a hand-arm teleportation system by combining a vision-based joint estimation approach, Transteleop, and an IMU-based arm teleoperation method.
    Transteleop aims to find identical kinematic features between the anthropomorphic robot hand and the human hand based on a human-to-robot translation model.
    The control connection between the hand and the arm is achieved by a self-designed camera holder, which makes this whole system mobile and less unrestricted.
    A series of robotic experiments, such as pushing a brick, dual-arm handover, and network evaluation on the test dataset verify the feasibility and reliability of our method.
    More implementation details, videos, and code are available at
\href{https://Smilels.github.io/multimodal-translation-teleop}{https://Smilels.github.io/multimodal-translation-teleop}.

    Although our method performs well in real-world tasks, it still has some limitations.
    First, the camera holder is an extra burden for the operator, which is not comfortable during long-term teleoperation.
    We would like to implement real-time hand tracking by a camera mounted on a robot arm to achieve an unlimited workspace for the novice. 
    Second, due to the lack of hand poses, which are commonly used in dexterous manipulation in our dataset, some high-precision tasks such as bottle opening and screw tightening are still intractable to the current system.
    Therefore, we plan to collect a hand dataset which focuses more on the subtle poses of the thumb, the first finger, and the middle finger.
    Third, slip detection and force estimation could be implemented to reduce the control burden on the user and to avoid unintentional collisions of the robot.
    
    \small{
        \section*{ACKNOWLEDGMENT}
        This research was funded jointly by the German Research Foundation (DFG) and the National Science Foundation of China (NSFC) in project Cross Modal Learning, NSFC 61621136008/DFG TRR-169. It was also partially supported by European project Ultracept (691154).
    }
    \bibliographystyle{IEEEtran} % (0.6 page, at least 25 refs)
    \bibliography{IEEEabrv,ref}
\end{document}